\documentclass[12pt]{article}
\usepackage[utf8]{inputenc}

\usepackage[margin=1in]{geometry}
\usepackage{amsfonts,amssymb,amsthm,amsmath,graphicx}
\usepackage{algorithm, algorithmic}
\usepackage{float}
\usepackage{listings}
\usepackage{hyperref}
\usepackage{verbatim}
\usepackage{mathtools}
\usepackage{enumitem}
\usepackage{xcolor}
\usepackage{caption}
\usepackage{multirow}
\hypersetup{
    colorlinks,
    linkcolor={red!50!black},
    citecolor={blue!50!black},
    urlcolor={blue!80!black}
}

\newcommand{\R}{\mathbb{R}}

\newcommand{\Exp}[1]{\mathbb{E}\left[ #1 \right]}

\definecolor{boxbackground}{rgb}{1,0.976, 0.882}

\title{AdaOja: Adaptive Learning Rates for Streaming PCA\thanks{NSF CAREER grant \#1255631}}
\author{Amelia Henriksen\thanks{Oden Institute, UT Austin, amelia@ices.utexas.edu} and Rachel Ward\thanks{Oden Institute, UT Austin, rward@math.utexas.edu}}

\begin{document}
\maketitle

\begin{abstract} 
Oja's algorithm has been the cornerstone of streaming methods in Principal Component Analysis (PCA) since it was first proposed in 1982. 
However, Oja's algorithm does not have a standardized choice of learning rate (step size) that both performs well in practice and truly conforms to the online streaming setting. In this paper, we propose a new learning rate scheme for Oja's method called AdaOja. 
This new algorithm requires only a single pass over the data and does not depend on knowing properties of the data set a priori.
 AdaOja is a novel variation of the Adagrad algorithm to Oja's algorithm in the single eigenvector case and extended to the multiple eigenvector case. 
 We demonstrate for dense synthetic data, sparse real-world data and dense real-world data that AdaOja outperforms common learning rate choices for Oja's method.
 We also show that AdaOja performs comparably to state-of-the-art algorithms (History PCA and Streaming Power Method) in the same streaming PCA setting.

\end{abstract}

\section{Introduction}
Modern problems give rise to more and more massive data sets.
Modern solutions require computationally efficient algorithms that make data manageable and therefore meaningful.
The main purpose of Principal Component Analysis is to project a data set with high dimension onto a data set of low dimension and still retain its fundamental properties.
Therefore, Principal Component Analysis is most useful for data sets that have such massive dimensions $n$ and $d$ that they cannot be stored or manipulated in practice.
Unfortunately, the prohibitive size of such data means that standard techniques for computing the principal components are also inefficient or impossible in practice.

This motivates the streaming setting for PCA, where the PCA algorithm iterates over the data set a few samples at a time to produce a basis for the desired lower-dimensional subspace of the data.
Of particular interest are algorithms for single-pass streaming PCA, which require only a single pass over the data set to obtain the desired subspace.
These algorithms are particularly important in the online setting, where data may only be read once.

Oja's method \cite{1982Oja} has been the fundamental basis for most streaming PCA results since its proposal in 1982.
This is largely because of its simplicity and asymptotic convergence guarantees under mild conditions \cite{1982Oja}.
However, Oja's method is commonly implemented using learning rate (step size) schemes that scale with an unknown constant.
This hyper-parameter must be predetermined to obtain optimal convergence rates--often requiring multiple passes over the data and violating the online setting.
Indeed, one of the fundamental problems with many streaming PCA algorithms is optimizing hyper-parameters without violating the streaming or the online settings.
In response to this deficiency we propose AdaOja, a new learning rate scheme for Oja's method for streaming PCA.
This method uses an adaptive scheme that circumvents the need to select or test over any hyper-parameters.
It is simple to implement and provides excellent convergence results in practice.

In section \ref{Sec: problem setting} we explain the problem setting and give background for Oja's  method.
In section \ref{Sec: AdaOja} we present the AdaOja algorithm.
In section \ref{Sec: Exp Results} we demonstrate that AdaOja performs as well as or better than multiple pass learning rates for Oja's method on both synthetic and real-world data.
This section demonstrates compelling empirical evidence for AdaOja's efficacy as a new implementation for Oja's method.
In section \ref{Sec: State of the Art}, we further demonstrate AdaOja's performance by comparing it to other state-of-the-art single-pass streaming algorithms across this variety of data sets.
Finally, we present future directions for our work.

\section{Problem Setting}
\label{Sec: problem setting}
Let $X \in \R^{n \times d}$ be a data set consisting of $n$ samples $x_t \in \R^{d}$.
We assume that these samples are i.i.d. with mean 0 and some unknown covariance $\Exp{x_tx_t^T} = \Sigma \in \R^{d \times d}$.
We want to find the $k$-dimensional, orthonormal subspace $W \in \R^{d \times k}$ s.t. when the data is projected onto $W$ the variance is maximized.
In other words, we want to solve
\begin{equation}
\max_{\substack{W \in \R^{d \times k}\\W^TW = I_k}} \text{tr}\left(W^T \Exp{x_tx_t^T} W \right) \equiv \min_{\substack{W \in \R^{d \times k}\\W^TW = I_k}}  -\text{tr}\left(W^T\Sigma W\right)
\end{equation}

Clearly, this equation is maximized by the top $k$ eigenvectors of $\Sigma$.
Since the true covariance matrix $\Sigma$ is unknown, classical PCA computes the top $k$ eigenvectors for the sample covariance matrix $\hat{\Sigma} = \frac{1}{n}\sum_{i=1}^nx_ix_i^T = \frac{1}{n}X^TX$.
Using the eigenvectors of the sample covariance matrix to approximate the eigenvectors of $\Sigma$ achieves the information lower bound \cite{2018Li, 2013Vu}.
As we mentioned before, however, computing the sample covariance and its corresponding eigenvectors directly using offline methods may be impossible for large $d$.

We recall that $\mathbb{E}[x_tx_t^TW] = \Sigma W \ \forall t \in [n]$. 
The gradient of the subspace $W$ in $-W^T\Sigma W$ is $-2\Sigma W$.
It follows that $x_tx_t^TW$ is an unbiased stochastic estimate of the gradient of our problem.

\subsection{Oja's Method}
\label{Section:Oja's Method}
The natural next step for this problem is to apply projected stochastic gradient descent (SGD), which is precisely Oja's method \cite{1982Oja}.
In the case that $k = 1$ we apply projected SGD as follows
\begin{enumerate}
    \item Initialize a vector of unit norm, $w_0$.
    \item For each $t$ in $[n]$ 
    \begin{enumerate}
        \item  Choose learning rate $\eta_t$
    	\item Set $w_t = (I + \eta_t x_tx_t^T)w_{t-1}$.
    	\item Project $w_t$ onto $S^{d-1}$ by taking $w_t = \frac{w_t}{||w_t||}$.
    \end{enumerate}
\end{enumerate}
A simple extension to the $k > 1$ case yields the following steps:
\begin{enumerate}
    \item Initialize a set of orthonormal vectors, $W \in \R^{d \times k}, W^TW = I_k$.
    \item For each $t$ in $[n]$
    \begin{enumerate}
        \item Choose learning rate $\eta_t$
    	\item Set $W_t = (I + \eta_t x_tx_t^T)W_{t-1}$.
    	\item Project $W_t$ onto $\{M \in \R^{d \times k}, M^TM = I_k\}$ by taking $W_t = Q_t R_t, W_t \leftarrow Q_t$ where $Q_tR_t$ is the QR-decomposition of $W_t$.
    \end{enumerate}
\end{enumerate}

\subsubsection{Learning Rates}
In many implementations of Oja's algorithm, it is common practice to choose $\eta_t = \frac{c}{t}$ or $\frac{c}{\sqrt{t}}$ where $c$ is a constant chosen by running the algorithm multiple times over the data.
For example, \cite{2018Balzano}, \cite{2018Xu}, \cite{2018Chen} and \cite{2016Li} implement Oja's with step size $\eta_t = c, c \in \R$; \cite{2018Yang},  \cite{2015Shamir} and \cite{2018Alakkari} use step size $\eta_t = \frac{c}{t}, c \in \R$ ; and \cite{2018Marinov} and \cite{2016Song} use step size  $\eta_t = \frac{c}{\sqrt{t}}, c \in \R$.
In these applications, typically many different possible constant $c$ values are tested on the data set, and only the best case values are used for empirical results.
This demonstrates the behavior of Oja's method in the best case, but this best case multi-pass PCA is not feasible for relevant online applications.
 Note that, even if multiple learning rates are applied in parallel (which would increase the memory constraints), determining the best case subspace to use requires taking and comparing accuracy readings for the all of the final solutions.
 This would be prohibitive for relevant accuracy metrics.
 Hence our goal is to find a robust learning rate scheme for Oja's method that does not require multiple passes, a priori information or wasteful parallelization.

A few papers suggest learning rate schemes distinct from the $c, \frac{c}{t}$ and $\frac{c}{\sqrt{t}}$ settings. 
For example, \cite{2006Lv} establishes an adaptive learning rate $\eta_t = \frac{\xi}{w_t^TX_t^TX_tw_t}$ for the single eigenvector case.
However, this method still includes hyperparameter $0 \leq \xi \leq 0.8$, the paper includes limited empirical results and \cite{2018Cardot} found that the method had poor performance.
In \cite{2018Chretien} the authors propose a complicated burn-in scheme followed by a probabilistically chosen step size scheme. 
The authors have not yet published many empirical results to justify their work, and the majority of the paper is theoretical.

Recent papers also suggest learning rate schemes for Oja's method based on theoretical guarantees \cite{2016Zhu, 2016Jain}, but these depend on parameters determined by the data set that we may not have a priori, such as the top eigenvalues of the covariance matrix. 
This again violates the streaming setting. 

\section{The AdaOja Algorithm}
\label{Sec: AdaOja}
\subsection{Background}
\subsubsection{AdaGrad}
We wanted to design a simple, effective way to determine the step size parameter for Oja's method without the need to fine tune hyperparameters or pre-determine properties of the data set.
This led us to consider common variants of stochastic gradient descent.
In 2010 \cite{2010McMahan} and \cite{2011Duchi} introduced the AdaGrad update step for stochastic gradient descent for a single vector.
This method is widely used in practice.
In \cite{2018Ward} the authors develop theory for a global step size variant of the AdaGrad update step.

For this scheme, the learning rate is defined via
\begin{equation} b_{t+1}^2 = b_t^2 + ||G_t||_2^2 \label{Equation:AdaGrad1} \end{equation}
\begin{equation}\eta_{t+1} = \frac{1}{b_{t+1}} \label{Equation:AdaGrad2} \end{equation}
\begin{equation}w_{t+1} \leftarrow w_t + \eta_{t+1}G_t \label{Equation:AdaGrad3} \end{equation}
Hence $\eta_t = \frac{1}{b_t}$.
Here $G_t \in \R^{d}$ is the latest stochastic approximation to the gradient.
Not only does this scheme work well when applied to SGD in practice, \cite{2018Ward} develops novel theory for this AdaGrad setting in non-convex landscapes.

We also considered how to adapt and apply other common learning rate schemes for stochastic gradient descent, such as ADAM \cite{2014Kingma} and RMSProp \cite{2012Hinton}. 
Both of these algorithms use a momentum term to improve the convergence rates for stochastic gradient descent.
However, \cite{2017DeSa} discusses why adding momentum naively to Oja's method fails in practice.
Indeed, when we implemented these methods and applied them to Oja's method, they completely failed as expected.
However, an interesting problem would be to see if a more sophisticated adaptation of ADAM or RMSProp could improve the results for Oja's method in practice.

\subsubsection{Mini-batching}
Note that a simple adaptation of the streaming PCA setting updates $W$ with mini-batches of samples $X_t \in \R^{B \times d}$ (where $B$ is small) rather than updating $W$ with a single sample $x_t \in \R^{d}$.
That is, for a mini-batch of samples $X_t \in \R^{B \times d}$ our stochastic approximation to the gradient would be $\frac{1}{B}X_t^TX_tW$ rather than $x_tx_t^TW$.
When $B = 1$ this is consistent with the single sample case.
Examples of this technique are found in \cite{2018Xu}, \cite{2018Yang}, \cite{2013Mitliagkas} and \cite{2014Hardt}.
This is particularly relevant for the $k > 1$ case where the time cost is dominated by the $QR$ decomposition of $W_t$. 
Manipulating the samples as a mini-batch therefore reduces the time complexity by a factor of $\frac{1}{B}$.
We apply this mini-batch scheme to our empirical implementations of Oja's method as well as our new algorithm, AdaOja, in Section \ref{Section:AdaOja}.

\subsection{AdaOja}
\label{Section:AdaOja}

We apply the adaptive learning rate method from Algorithm 2 in \cite{2018Ward} to Oja's method \cite{1982Oja} to obtain the AdaOja algorithm.
To the best of our knowledge, AdaGrad has never before been applied to Oja's method and the streaming PCA problem in this way. 

\begin{algorithm}
\caption{AdaOja, $k=1$ case}
\begin{algorithmic}
\STATE{input: $X_1, \ldots, X_N$, $X_t \in \R^{B \times d}, b_0 \approx 10^{-5}$}
\STATE{$w_0 \sim [N(0,1)]^k$}
\STATE{$w_0 \gets \frac{w_0}{|| w_0||_2}$}
\FOR{$t \in 1, \ldots, N$}
    \STATE{ $G_t \gets \frac{1}{B}X_t^TX_tw_{t-1}$}
    \STATE{ $b_t \gets \sqrt{b_{t-1}^2 +||G_t ||_2^2}$}
    \STATE{ $w_t \gets w_{t-1} + \frac{1}{b_t} G_t$}
    \STATE{$w_t \gets \frac{w_t}{|| w_t||_2}$}
\ENDFOR
\end{algorithmic}
\label{Alg:AdaOja k=1}
\end{algorithm}
Note that \cite{2018Ward} establishes that AdaGrad (equations \ref{Equation:AdaGrad1}-\ref{Equation:AdaGrad3}) is strongly robust to the initial choice of $b_0$.
We found $b_0 \approx 10^{-5}$ to be a sufficiently small starting size based on the empirical results of \cite{2018Ward} and our own work.

We want to extend the $k=1$ case to the $k>1$ case.
Yet the global step size AdaGrad algorithm from equations \ref{Equation:AdaGrad1}-\ref{Equation:AdaGrad3} assumes $w_t, G_t \in \R^d$.
The simplest extension of AdaGrad to the $k>1$ case updates the learning rate with the squared $L^2$ norm of the entire matrix $G_t \in \R^{d \times k}$.
A better extension, however, draws on this AdaGrad algorithm in its coordinate form.
In the $k=1$, coordinate form, each of the $d$ components receives and updates its own learning rate, setting: \begin{equation} b_{t+1}^2[i] = b_t^2[i] + G_t[i]^2.\end{equation}
We can apply this principle to the $k>1$ case by obtaining and updating unique learning rates for each of the $k$ \emph{columns} of $W_t$.
We have chosen not to use the full coordinate form of AdaGrad to avoid over-constraining the problem and find that this column-wise extension works well in practice.
Algorithm \ref{Alg:AdaOja k>1} is the result.

\begin{algorithm}
\caption{AdaOja, $k>1$ case, vectorized $b_t$}
\begin{algorithmic}
\STATE{input: $X_1, \ldots, X_N$, $X_t \in \R^{B \times d}, b_0 \approx 10^{-5}$}
\STATE{$Q_0 \sim [N(0,1)]^{d\times k}$}
\STATE{$Q_0 \gets QR[Q_0][0]$}
\FOR{$t \in 1, \ldots, N$}
    \STATE{$G_t \gets \frac{1}{B}X_t^TX_tQ^{(t-1)}$}
    \FORALL{$i \in 1, \ldots, k$}
    	\STATE{$b_t[i] \gets \sqrt{b_{t-1}[i]^2 + ||G_t[:,i]||_2^2}$}
        \STATE{$Q_{t}[:,i] \gets Q_{(t-1)}[:,i] + \frac{1}{b_t[i]} G_t[:,i]$}
    \ENDFOR
    \STATE{$Q_t \gets QR[Q_t][0]$}
\ENDFOR
\end{algorithmic}
\label{Alg:AdaOja k>1}
\end{algorithm}

\section{Testing and Empirical Results}
\label{Sec: Exp Results}
\subsubsection{Accuracy Metric}
The PCA problem seeks to find the directions of maximum variation for our data set.
That is, when we project our data set onto a subspace of lower dimension, we want to capture the maximum amount of variance possible in the data set.
Hence, an obvious metric for the accuracy of our model is the explained variance, defined (for $\Exp{X} = 0$) to be: \begin{equation}\frac{tr(W^TX^TXW)}{||X||_F^2} = \frac{||XW||_F^2}{||X||_F^2}.\label{eq: explained variance} \end{equation}
This metric is essentially the percentage of the original variance of the data set captured by our new, lower dimensional data set.
Note that the explained variance is maximized by the eigenvectors of of the sample covariance $\hat{\Sigma}$.
The explained variance is particularly useful for several reasons.
 
 First, the explained variance  demonstrates how much of the original data set can be represented in a lower dimension.
 When $W$ is equal to the top $k$ eigenvectors of the sample covariance matrix, the explained variance is the maximum amount of variance we can recover given the data we have received.
Second, the explained variance is directly connected to the Rayleigh quotient and subsequently the problem setting.
Third, the explained variance is robust to situations where the gap between the top eigenvalues is $\ll 1$. 
    That is, if the top two eigenvalues are the same (or nearly the same), then the eigenvector associated with either the first or second eigenvalue would sufficiently maximize the variance in the problem setting. 
    This is reflected in the explained variance.
    For error metrics that are concerned with retrieving the exact eigenvectors of $\hat{\Sigma}$ (such as the commonly used principal angle based error metric), interchanging the eigenvectors would cause the algorithm to fail to converge.

\subsection{The Power of AdaOja Learning Rates}

To test the efficacy of our algorithm, we test on both synthetic and real-world data.
Our experiments on synthetic data are included in section \ref{Section:SpCov}, our experiments on sparse real-world data are included in section \ref{Section:bag-of-words}, and our experiments on dense real-world data are included in section \ref{Section:CIFAR}.  
For each data set, we run AdaOja against Oja's method with learning rates $\frac{c}{t}, \frac{c}{\sqrt{t}}$ respectively. 
In our plots, we show the final explained variance achieved by Oja's method for $c$ on a range of scales, and plot these against the final explained variance achieved AdaOja (with $b_0 = 10^{-5}$) in a single pass.
For sufficiently small data sets, we also plot the final explained variance for the eigenvectors of the sample covariance computed explicitly in the offline setting.
Both this final "svd" derived value and the final AdaOja value are kept constant for all $c$ and serve as a reference.

We find that across every data type, AdaOja outperforms Oja's method for the majority of scales for $c$. 
It also achieves compellingly close explained variance results to the "true" offline explained variance results--particularly for our dense real-world data and low-noise synthetic data.

\subsubsection{Spiked Covariance Data}
\label{Section:SpCov}
For our experiments with synthetic data, we use the spiked covariance model.
In our implementation of this model, we let $x_t \in \R^{d}, x_t \sim N(\mathbf{0}, \Sigma)$ where $\Sigma \in \R^{d\times d}$
\begin{equation}\Sigma = A_0 Diag(w)^2 A_0^T + \sigma^2I\end{equation}
Here $A_0 \in \R^{d, k}$ is a set of $k$ d-dimensional orthonormal vectors.
We set $w \in \R^{k}$ to be s.t. $w_t \sim U(0,1)$ and $w_1 \geq w_2 \geq \ldots \geq w_k$.
We scale $w_t = \frac{w_t}{w_1} \ \forall t$ so that $w_1 = 1$ and set $Diag(w) \in \R^{k,k}$ to be the diagonal matrix with values $w_t$. 
Here $\sigma$ is a noise parameter that augments the distribution.
In our examples, we set $n=10000$ and $d = 1000$.

For this data, we measure the final explained variance for $c = 5^i \ \forall \ i \in \{-5, -4, \dots, 10\}$.
We ran these tests with batch size $B = 10$ for $k \in \{1, 5, 10\}$ and $\sigma \in \{.1, .25, .5, .75, 1\}$ to demonstrate the behavior over a range of values.
Here we include the results for $\sigma \in \{.1, .75\}, k \in \{1, 5, 10\}$ to demonstrate the differences in the high and low noise cases--see figures \ref{fig:spcov sigma=.1} and \ref{fig:spcov sigma=.75}.
The remaining plots are included in Appendix \ref{App: Additional Vis}.

\begin{figure}[H]
    \centering
    \includegraphics[scale=.5]{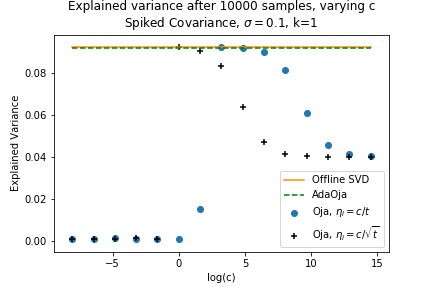}
    \includegraphics[scale=.5]{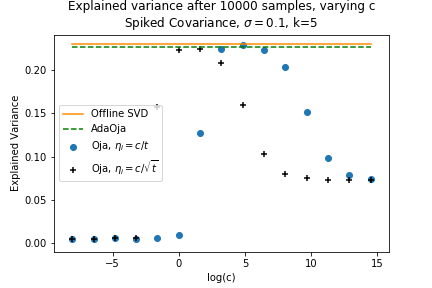}
    \includegraphics[scale=.5]{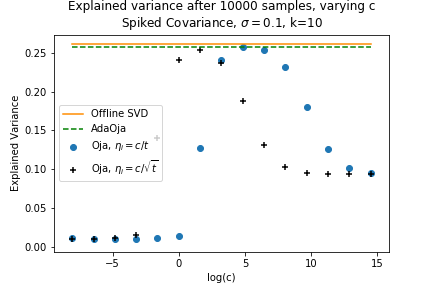}
    \caption{Spiked Covariance model for low noise $\sigma=0.01, k \in \{1, 5, 10\}, B=10, n = 10000, d = 1000$, plot of final explained variance for Oja's method with learning rate $\eta_t = \frac{c}{t}, \frac{c}{\sqrt{t}}, c = 5^{i} \ \forall i \in \{-5, -4, \cdots, 10\}$. Final explained variance achieved by AdaOja and by the offline principal components of the sample covariance matrix included for reference.}
    \label{fig:spcov sigma=.1}
\end{figure}

\begin{figure}[H]
    \centering
    \includegraphics[scale=.5]{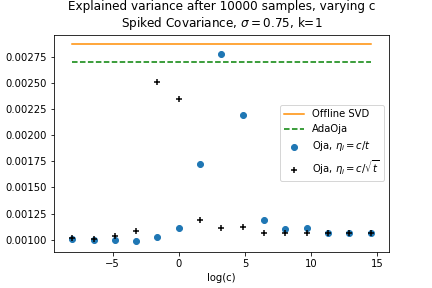}
    \includegraphics[scale=.5]{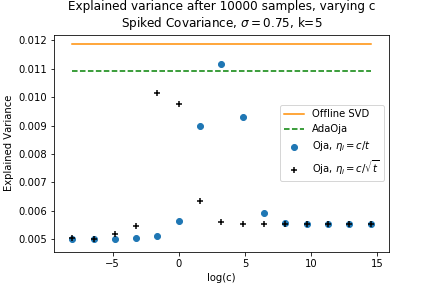}
    \includegraphics[scale=.5]{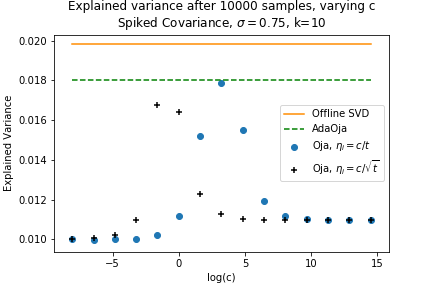}
    \caption{Spiked Covariance model for high noise $\sigma=0.75, k \in \{1, 5, 10\}, B=10, n = 10000, d = 1000$, plot of final explained variance for Oja's method with learning rate $\eta_t = \frac{c}{t}, \frac{c}{\sqrt{t}}, c = 5^{i} \ \forall i \in \{-5, -4, \cdots, 10\}$. Final explained variance achieved by AdaOja and by the offline principal components of the sample covariance matrix included for reference.}
    \label{fig:spcov sigma=.75}
\end{figure}

From our synthetic data sets, we see certain trends. 
First, for instances of low noise, the final explained variance for Oja's method with both $\eta_t = \frac{c}{t}$ and $\eta_t = \frac{c}{\sqrt{t}}$ varies widely depending on the scaling of $c$. 
As the noise increases, there is less variation in the final explained variance regardless of the scaling of $c$.
We further notice that as $k$ increases, there is a greater difference between the maximum final explained variance achieved by Oja's and the minimum final explained variance achieved by Oja's.
This trend is consistent for every noise level $\sigma$.

We also note that the maximum explained variance achieved across $c$ values is approximately the same for both Oja's with learning rate $\eta_t = \frac{c}{t}$ and Oja's with learning rate $\eta_t = \frac{c}{\sqrt{t}}$.
It is most significant, however, that for every instance of $\sigma$ and $k$, the final explained variance of AdaOja approximately achieves or outperforms this maximum explained variance without any hyper-parameter optimization.
Not only does AdaOja outperform Oja's with learning rates $c/t$ and $c/\sqrt{t}$ for the vast majority of scales, it achieves the best results possible almost every time without violating the single-pass streaming setting to determine learning rate hyperparameters.
Note that for low noise, there is almost no difference between the explained variance achieved offline from the sample covariance and online via AdaOja. As the noise increases and as $k$ increases, there is a marginal gap between this explicit value and the value achieved with AdaOja.

\subsubsection{Sparse bag-of-words}
\label{Section:bag-of-words}
We apply our algorithm to  five different real world, sparse, bag-of-words data sets: Kos, NIPS, Enron, Nytimes, and PubMed \cite{2017Dua}.
All of these data sets are sparse, with densities ranging from 0.0004 to 0.04. 

\begin{figure}[H]
    \centering
    \includegraphics[scale=.5]{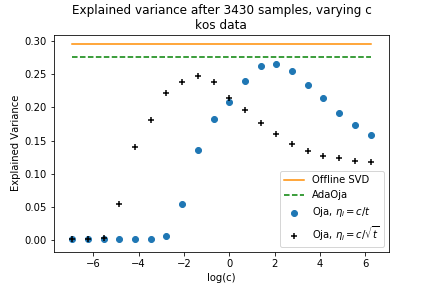} 
    \includegraphics[scale=.5]{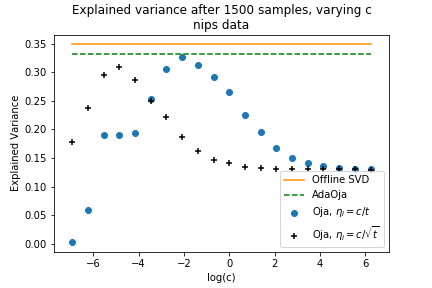} 
    \includegraphics[scale=.5]{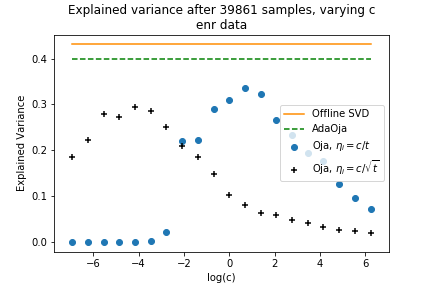}
    \caption{Real world, small bag-of-words data sets. Here $k=10, B=10, c=2^i \forall i \in \{-5, \cdots, 15\}$.Upper left: Kos Dataset (n = 3430, d=6906); upper right: Nips Dataset (n=1500, d=12419); bottom center: Enron Dataset (n=39861, d=28102).}
    \label{fig:small bag cvals}
\end{figure}

For our small bag-of-words data (Kos, NIPS, and Enron) we set $c = 2^i \ \forall \ i \in \{-10, \dots, 10\}$.
We run Oja's and AdaOja's with batch size $B = 10$ and seek to recover the top $k=10$ eigenvectors.
These results are visualized in figure \ref{fig:small bag cvals}.
For these data sets, AdaOja achieves greater explained variance than Oja's for both learning rates for every choice of $c$.
For these data sets, we also notice that the maximum explained variance for $\eta_t = \frac{c}{t}$ is slightly greater than the maximum explained variance for $\eta_t = \frac{c}{\sqrt{t}}$.
We also note that Oja's with $\eta_t = \frac{c}{t}$ and $\eta_t = \frac{c}{\sqrt{t}}$ achieved their best performance for a very limited number scales--without a priori information or multiple passes to determine $c$, it is unlikely that Oja's would perform well in these settings.

\begin{figure}[H]
    \centering
    \includegraphics[scale=.5]{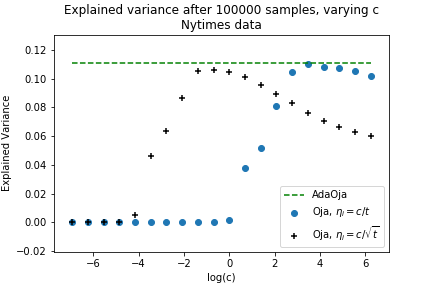}
    \includegraphics[scale=.5]{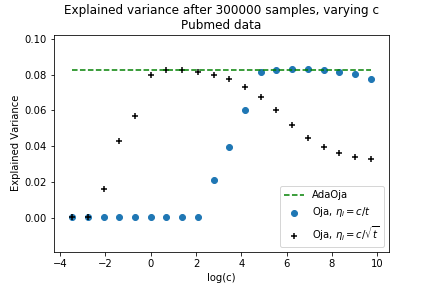}
    \caption{Real world, small bag-of-words data sets. Here $k=10, B=10, c=2^i \forall i \in \{-5, \cdots, 15\}$.Upper left: NyTimes Dataset (d=102,660), upper right: Pubmed dataset (d=141,041). Note that for the NyTimes data set we limited our results to the first 100,000 samples and for the PubMed data set, we limited our results to the first 300,000 samples. Due to the prohibitive size of these data sets, we do not compute the explained variance of the offline principal components for the sample covariance.}
    \label{fig:big bag cvals}
\end{figure}

For our NyTimes, we again test $c = 2^i \ \forall \ i \in \{-10, \dots, 10\}$ and for our Pubmed data set we choose $c=2^i \ \forall \ i \in \{-5, \dots, 15\}$.
For both of these we set the batch size to be $B = 100$ and seek to recover the top $k=10$ eigenvectors.
These results are visualized in figure \ref{fig:big bag cvals}.
In the case of these larger data sets, we note there are slightly more $c$ values for which Oja's achieves best case behavior, but the algorithm still falls short for the vast majority of $c$ values.
As with our previous results, we note that AdaOja's method achieves greater than or equal to the best case explained variance for Oja's method without any hyperparameter optimization. 

\subsubsection{CIFAR-10 Data set}
\label{Section:CIFAR}
\begin{figure}[H]
    \centering 
    \includegraphics[scale=.5]{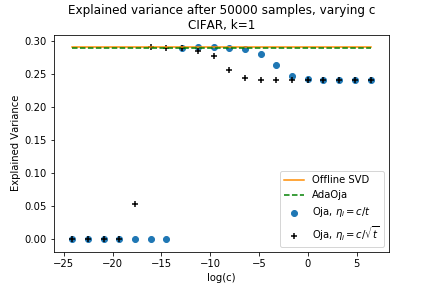}
    \includegraphics[scale=.5]{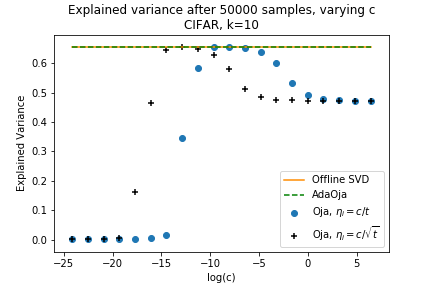}
    \caption{CIFAR-10 Data set, $k = \{1, 10\}$, $n=50000, \ d = 1024, \ B=10$, $c \in \{5^i | i \in -15, \dots, 5\}$}
    \label{fig:CIFAR cvals}
\end{figure}

CIFAR 10 is a subset of the tiny images data set \cite{2009Krizhevsky}.
It contains 50,000 training and 10,000 testing $32 \times 32$ color images in 10 mutually exclusive classes (6,000 images per class) and is frequently used for image classification.
Note that because this data is dense, we first centralized the data by subtracting the mean of each attribute (pixel) before applying our algorithm.
Figure \ref{fig:CIFAR cvals} exhibits the final explained variance after 50,000 samples for AdaOja vs Oja's method with learning rates $\eta_t = \frac{c}{t}, \ \eta_t = \frac{c}{\sqrt{t}}, \ c \in \{5^i|i\in -15, \dots, 5\}$.
As before, we see that the final explained variance achieved by Oja's method is significantly lower than that achieved by AdaOja for the majority of scales.
We also note that, as with our spiked covariance data, as $k$ increases the gap between the explained variance achieved by AdaOja and the explained variance achieved by Oja's increases for the majority of the scales.
We also note that AdaOja achieves almost exactly the same explained variance as the "true" principal components computed offline.

\section{AdaOja vs. State of the Art}
\label{Sec: State of the Art}
One of our objectives for this work is to not only demonstrate the ability of AdaOja's method to solve the learning rate problem for Oja's method, but to test the performance of AdaOja against other, state-of-the-art streaming solutions. 
In particular, we test the convergence of the explained variance for the AdaOja algorithm against two other algorithms: Streaming Power Method (SPM) and History PCA (HPCA).
Streaming Power Method (which is the noisy power method applied to the streaming PCA problem) was first introduced for PCA in \cite{2013Mitliagkas}.
Further theory for this method in the PCA setting was developed in \cite{2014Hardt} and \cite{2016Balcan}.
Yang, Hsieh, and Wang recently proposed a new algorithm \cite{2018Yang} for streaming PCA called History PCA (HPCA) which performs PCA in the block streaming setting using an update step that combines the block power method and Oja's algorithm.
This algorithm performs well in practice and does not require hyperparameter estimation for the learning rate.
It is interesting to note that in the empirical results from \cite{2018Yang}, Oja's method was implemented with $\eta_t = \frac{c}{t}$ for a range of $c$ values (with the top results displayed).
In these experiments, HPCA consistently--and sometimes significantly--outperformed Oja's method.
However, our results demonstrate that the adaptive nature of AdaOja appears to compensate for some of these deficiencies.

\begin{figure}[H]
    \centering
    \includegraphics[scale=.5]{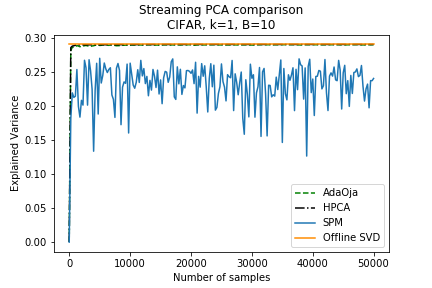}
    \includegraphics[scale=.5]{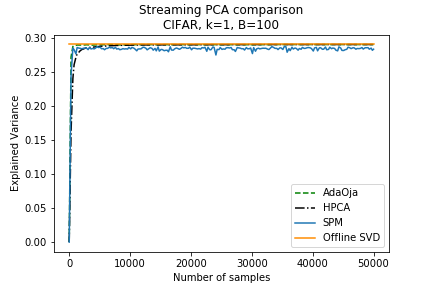}
    \includegraphics[scale=.5]{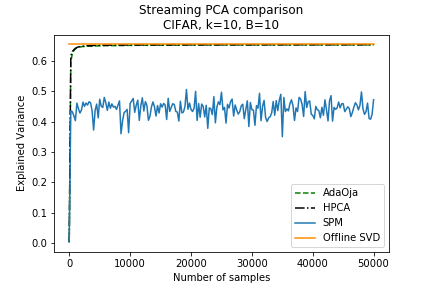}
    \includegraphics[scale=.5]{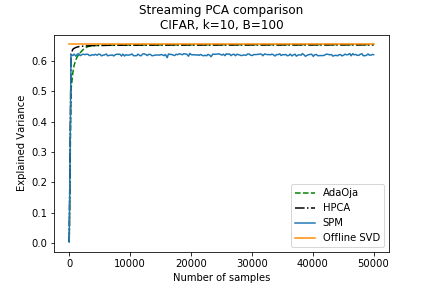}
    \caption{Explained variance plotted against the sample number for the real-world CIFAR dataset. Here $k \in \{1, 10\}, B \in \{10, 100\}$. Note that the final explained variance for the offline computed principal components of the sample covariance is included here for reference.}
    \label{fig:expvarcomp CIFAR}. 
\end{figure}

In our experiments, we tested the explained variance for AdaOja, HPCA, and SPM on the data sets from section \ref{Sec: Exp Results} for both small and large batch sizes $B$.
Varying the batch size is important to demonstrate the behavior of SPM, which as a stochastic variant of the standard power method is highly dependent on the choice of $B$.
For example, for the real-world, dense CIFAR data set, AdaOja and HPCA achieve almost identical convergence in both the $B=10$ and $B=100$ settings. 
Yet SPM only achieves comparable convergence with these methods in the $B=100$ setting, and for $k=10$ still falls short of the near optimal explained variance achieved by AdaOja and HPCA.
Figure \ref{fig:expvarcomp CIFAR} demonstrates this result.

\begin{figure}[H]
    \centering
    \includegraphics[scale=.5]{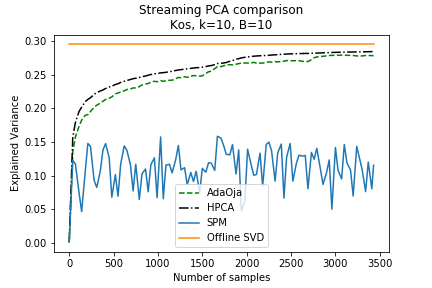}
    \includegraphics[scale=.5]{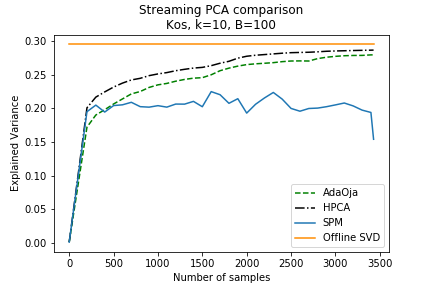}
    \includegraphics[scale=.5]{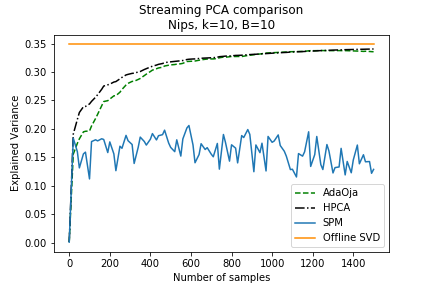}
    \includegraphics[scale=.5]{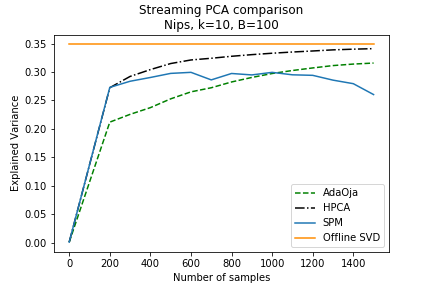}
    \includegraphics[scale=.5]{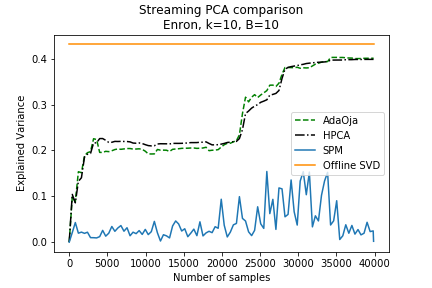}
    \includegraphics[scale=.5]{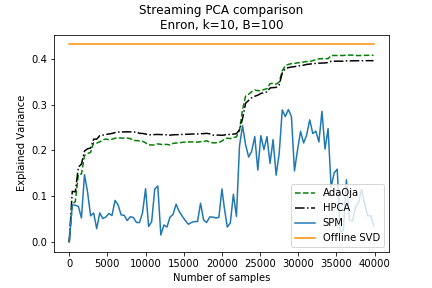}
    \caption{Explained variance plotted against the sample number for the small, real-world, sparse Bag-of-Words data sets: Kos, Nips and Enron. Here $k = 10, B \in \{10, 100\}$ CIFAR dataset.}
    \label{fig:expvarcomp small bag}
\end{figure}

\begin{figure}[H]
    \centering
    \includegraphics[scale=.5]{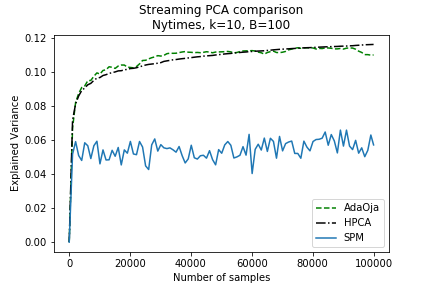}
    \includegraphics[scale=.5]{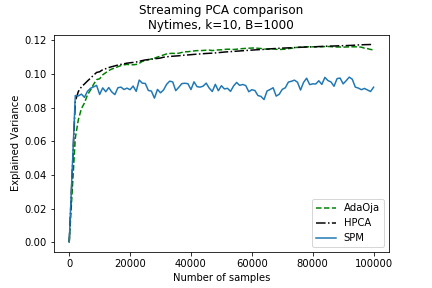}
    \includegraphics[scale=.5]{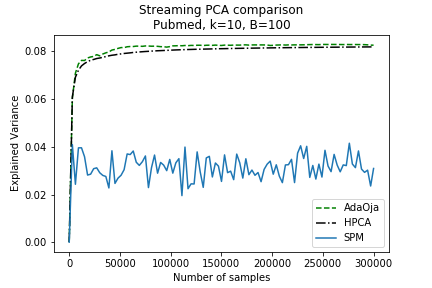}
    \includegraphics[scale=.5]{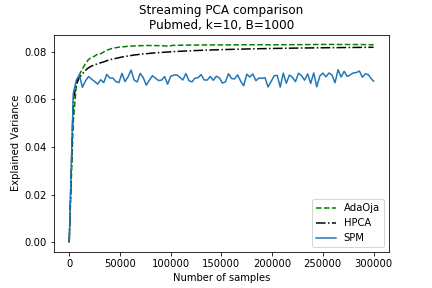}
    \caption{Explained variance plotted against the sample number for the medium, real-world, sparse Bag-of-Words data sets: Nytimes and Pubmed. Here $k = 10, B \in \{100, 1000\}$}
    \label{fig:expvarcomp big bag}
\end{figure}

AdaOja appears to perform comparably to HPCA, and in some instances outperforms it. 
For example, figure \ref{fig:expvarcomp small bag} compares the convergence of the three methods for our small bag-of-words data sets, on which HPCA either marginally outperforms AdaOja (as with the Kos and Nips data sets) or performs approximately the same (as with the Enron data set). 
However, for our slightly larger Bag-of-Words data sets (see figure \ref{fig:expvarcomp big bag}), AdaOja appears to marginally outperform HPCA.
Hence the adaptive choice of stepsize enables Oja's method to legitimately compete with state-of-the-art methods for real-world data.

We notice some of the same trends on our synthetic spiked covariance data. 
In particular, the performance of AdaOja and HPCA appears to be fairly consistent for $B=10$ and $B=100$ (with AdaOja sometimes achieving improvement in the $B=100$ case), but SPM achieves far better explained variance in the $B=100$ case.
For higher noise levels, however, SPM achieves worse and worse explained variance with only marginal improvements in the $B=100$ case.
We also note that across all choices of $k$, AdaOja tends to achieve better convergence rates and a higher explained variance than HPCA as the noise increases.
The images for these results are contained in Appendix \ref{App:SotA comparison}.

\section{Future Work}
In this paper we introduced AdaOja, a new algorithm for streaming PCA based on Oja's method and a global stepsize variant of the AdaGrad algorithm. 
This algorithm provides a simple solution to the hyperparameter optimization problem for Oja's method that is easy to implement and works well in practice.
We demonstrated on multiple different types of data that this algorithm approximately achieves or surpasses the optimal performance of Oja's method against other commonly used learning rate schemes.
We also showed that this algorithm performs comparably to or surpasses other state-of-the-art algorithms, and is robust to the choice of batch size (unlike SPM).

These compelling empirical results open intriguing new avenues for research.
Several algorithms for streaming PCA incorporate Oja's method into their update steps. 
For example, HPCA \cite{2018Yang} uses an update step that combines Oja's method with the block power method. 
One interesting area of further development would be to incorporate AdaOja--rather than Oja's method--into this algorithm.
Another algorithm, Oja++ \cite{2016Zhu} suggests a gradual initialization period after which the algorithm proceeds exactly as Oja's method. 
It may yield better results to incorporate AdaOja into this scheme.

Of course, given these compelling empirical convergence results we want to establish theoretical convergence guarantees for AdaOja as well.
This is a particularly exciting field of research because there is an extraordinary gap between methods implemented in theory and in practice for such iterative methods.
Our method is essentially a variant of stochastic gradient descent, projected onto the (nm-convex) unit sphere.
However, the setting is non-convex, the learning rate is adaptive, and the result at each iteration is projected onto the space of orthogonal unit vectors.
Theoretical results for AdaGrad in the stochastic, non-convex setting are only recently being developed and only guarantee convergence to a critical point \cite{2018Ward, 2018Zou}. 
Theoretical results for Oja's method are also an open area of research, and current convergence rates have been largely derived from complex learning rate schemes that are neither practically usable nor applicable in the adaptive setting (see \cite{2016Jain, 2016Zhu} for some of the most recent work in this area).
Hence, establishing convergence rate results for AdaOja will lead to novel theoretical improvements for both AdaGrad and Oja's method.

\bibliographystyle{plain}
\bibliography{Bibliography}

\begin{thebibliography}{10}

\bibitem{2018Alakkari}
Salaheddin Alakkari and John Dingliana.
\newblock {An Acceleration Scheme for Memory Limited, Streaming PCA}.
\newblock {\em ArXiv e-prints}, July 2018.

\bibitem{2016Zhu}
Z.~{Allen-Zhu} and Y.~{Li}.
\newblock {First Efficient Convergence for Streaming k-PCA: a Global, Gap-Free,
  and Near-Optimal Rate}.
\newblock {\em ArXiv e-prints}, July 2016.

\bibitem{2018Balzano}
L.~Balzano, Y.~Chi, and Y.~M. Lu.
\newblock Streaming pca and subspace tracking: The missing data case.
\newblock {\em Proceedings of the IEEE}, 106(8):1293--1310, Aug 2018.

\bibitem{2018Cardot}
Hervé Cardot and David Degras.
\newblock Online principal component analysis in high dimension: Which
  algorithm to choose?
\newblock {\em International Statistical Review}, 86(1):29--50, 2018.

\bibitem{2018Chen}
Minshuo Chen, Lin Yang, Mengdi Wang, and Tuo Zhao.
\newblock Dimensionality reduction for stationary time series via stochastic
  nonconvex optimization.
\newblock {\em CoRR}, abs/1803.02312, 2018.

\bibitem{2018Chretien}
Stephane {Chretien}, Christophe {Guyeux}, and Zhen-Wai {Olivier HO}.
\newblock {Average performance analysis of the stochastic gradient method for
  online PCA}.
\newblock {\em arXiv e-prints}, page arXiv:1804.01071, Apr 2018.

\bibitem{2017DeSa}
Christopher {De Sa}, Bryan {He}, Ioannis {Mitliagkas}, Christopher {R{\'e}},
  and Peng {Xu}.
\newblock {Accelerated Stochastic Power Iteration}.
\newblock {\em arXiv e-prints}, page arXiv:1707.02670, Jul 2017.

\bibitem{2017Dua}
Dheeru Dua and Efi Karra~Taniskidou.
\newblock {UCI} machine learning repository, 2017.

\bibitem{2011Duchi}
John Duchi, Elad Hazan, and Yoram Singer.
\newblock Adaptive subgradient methods for online learning and stochastic
  optimization.
\newblock {\em J. Mach. Learn. Res.}, 12:2121--2159, July 2011.

\bibitem{2016Balcan}
Maria {Florina Balcan}, Simon~S. {Du}, Yining {Wang}, and Adams~Wei {Yu}.
\newblock {An Improved Gap-Dependency Analysis of the Noisy Power Method}.
\newblock {\em arXiv e-prints}, page arXiv:1602.07046, Feb 2016.

\bibitem{2014Hardt}
Moritz Hardt and Eric Price.
\newblock The noisy power method: A meta algorithm with applications.
\newblock In {\em Proceedings of the 27th International Conference on Neural
  Information Processing Systems - Volume 2}, NIPS'14, pages 2861--2869,
  Cambridge, MA, USA, 2014. MIT Press.

\bibitem{2012Hinton}
G.~Hinton, N.~Srivastava, and K.~Swersky.
\newblock Lecture 6a overview of mini-batch gradient descent".
\newblock Coursera Lectures slides, 2012.
\newblock Available at:
  https://www.cs.toronto.edu/~tijmen/csc321/slides/lecture\_slides\_lec6.pdf.

\bibitem{2016Jain}
Prateek Jain, Chi Jin, Sham~M. Kakade, Praneeth Netrapalli, and Aaron Sidford.
\newblock Matching matrix bernstein with little memory: Near-optimal finite
  sample guarantees for oja's algorithm.
\newblock {\em CoRR}, abs/1602.06929, 2016.

\bibitem{2016Li}
Chris {Junchi Li}, Mengdi {Wang}, Han {Liu}, and Tong {Zhang}.
\newblock {Near-Optimal Stochastic Approximation for Online Principal Component
  Estimation}.
\newblock {\em arXiv e-prints}, page arXiv:1603.05305, Mar 2016.

\bibitem{2014Kingma}
Diederik~P. Kingma and Jimmy Ba.
\newblock Adam: {A} method for stochastic optimization.
\newblock {\em CoRR}, abs/1412.6980, 2014.

\bibitem{2009Krizhevsky}
Alex Krizhevsky.
\newblock Learning multiple layers of features from tiny images.
\newblock 2009.

\bibitem{2018Li}
Chris~Junchi Li, Mengdi Wang, Han Liu, and Tong Zhang.
\newblock Near-optimal stochastic approximation for online principal component
  estimation.
\newblock {\em Mathematical Programming}, 167(1):75--97, Jan 2018.

\bibitem{2006Lv}
Jian~Cheng Lv, Zhang Yi, and K.K. Tan.
\newblock Global convergence of oja's pca learning algorithm with a
  non-zero-approaching adaptive learning rate.
\newblock {\em Theoretical Computer Science}, 367(3):286 -- 307, 2006.

\bibitem{2018Marinov}
Teodor~Vanislavov Marinov, Poorya Mianjy, and Raman Arora.
\newblock Streaming principal component analysis in noisy setting.
\newblock In Jennifer Dy and Andreas Krause, editors, {\em Proceedings of the
  35th International Conference on Machine Learning}, volume~80 of {\em
  Proceedings of Machine Learning Research}, pages 3413--3422,
  Stockholmsmässan, Stockholm Sweden, 10--15 Jul 2018. PMLR.

\bibitem{2010McMahan}
H.~Brendan McMahan and Matthew~J. Streeter.
\newblock Adaptive bound optimization for online convex optimization.
\newblock {\em CoRR}, abs/1002.4908, 2010.

\bibitem{2013Mitliagkas}
I.~{Mitliagkas}, C.~{Caramanis}, and P.~{Jain}.
\newblock {Memory Limited, Streaming PCA}.
\newblock {\em ArXiv e-prints}, June 2013.

\bibitem{1982Oja}
Erkki Oja.
\newblock Simplified neuron model as a principal component analyzer.
\newblock {\em Journal of Mathematical Biology}, 15(3):267--273, Nov 1982.

\bibitem{2015Shamir}
Ohad Shamir.
\newblock A stochastic pca and svd algorithm with an exponential convergence
  rate.
\newblock In {\em ICML}, 2015.

\bibitem{2016Song}
W.~Song, J.~Zhu, Y.~Li, and C.~Chen.
\newblock Image alignment by online robust pca via stochastic gradient descent.
\newblock {\em IEEE Transactions on Circuits and Systems for Video Technology},
  26(7):1241--1250, July 2016.

\bibitem{2013Vu}
Vincent~Q. Vu and Jing Lei.
\newblock Minimax sparse principal subspace estimation in high dimensions.
\newblock {\em Ann. Statist.}, 41(6):2905--2947, 12 2013.

\bibitem{2018Ward}
Rachel {Ward}, Xiaoxia {Wu}, and Leon {Bottou}.
\newblock {AdaGrad stepsizes: Sharp convergence over nonconvex landscapes, from
  any initialization}.
\newblock {\em arXiv e-prints}, page arXiv:1806.01811, Jun 2018.

\bibitem{2018Xu}
Peng Xu, Bryan He, Christopher~De Sa, Ioannis Mitliagkas, and Chris Re.
\newblock Accelerated stochastic power iteration.
\newblock In Amos Storkey and Fernando Perez-Cruz, editors, {\em Proceedings of
  the Twenty-First International Conference on Artificial Intelligence and
  Statistics}, volume~84 of {\em Proceedings of Machine Learning Research},
  pages 58--67, Playa Blanca, Lanzarote, Canary Islands, 09--11 Apr 2018. PMLR.

\bibitem{2018Yang}
P.~{Yang}, C.-J. {Hsieh}, and J.-L. {Wang}.
\newblock {History PCA: A New Algorithm for Streaming PCA}.
\newblock {\em ArXiv e-prints}, February 2018.

\bibitem{2018Zou}
Fangyu Zou and Li~Shen.
\newblock On the convergence of adagrad with momentum for training deep neural
  networks.
\newblock {\em CoRR}, abs/1808.03408, 2018.

\end{thebibliography}

\appendix

\section{Additional Visualizations for AdaOja vs. Oja}
\label{App: Additional Vis}
\begin{figure}[H]
    \centering
    \includegraphics[scale=.5]{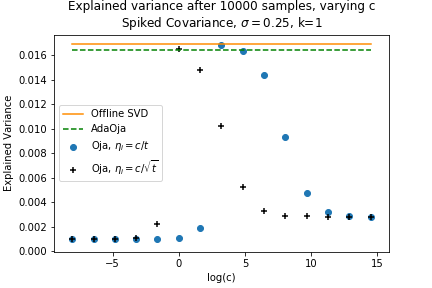}
    \includegraphics[scale=.5]{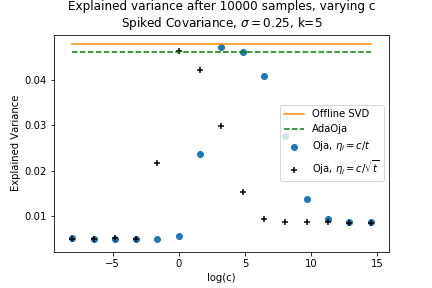}
    \includegraphics[scale=.5]{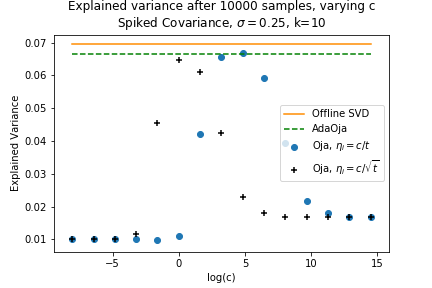}
    \caption{Spiked Covariance model for noise $\sigma=0.25, k \in \{1, 5, 10\}, B=10, n = 10000, d = 1000$, plot of final explained variance for Oja's method with learning rate $\eta_t = \frac{c}{t}, \frac{c}{\sqrt{t}}, c = 5^{i} \ \forall i \in \{-5, -4, \cdots, 10\}$.}
\end{figure}

\begin{figure}[H]
    \centering
    \includegraphics[scale=.5]{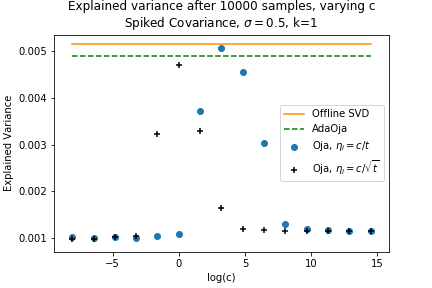}
    \includegraphics[scale=.5]{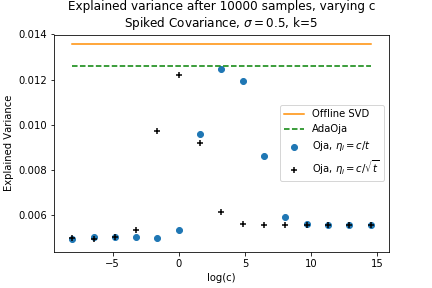}
    \includegraphics[scale=.5]{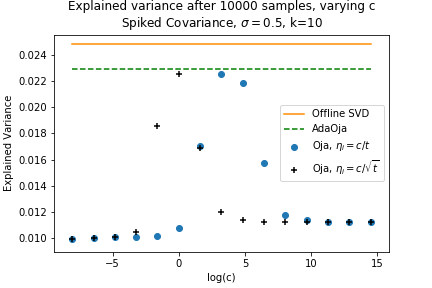}
    \caption{Spiked Covariance model for noise $\sigma=0.5, k \in \{1, 5, 10\}, B=10, n = 10000, d = 1000$, plot of final explained variance for Oja's method with learning rate $\eta_t = \frac{c}{t}, \frac{c}{\sqrt{t}}, c = 5^{i} \ \forall i \in \{-5, -4, \cdots, 10\}$.}
\end{figure}

\begin{figure}[H]
    \centering
    \includegraphics[scale=.5]{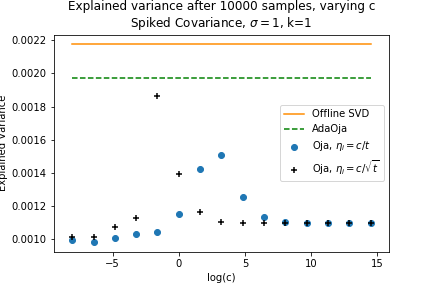}
    \includegraphics[scale=.5]{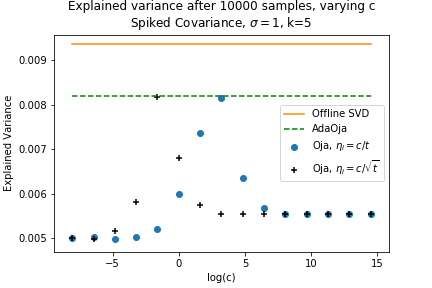}
    \includegraphics[scale=.5]{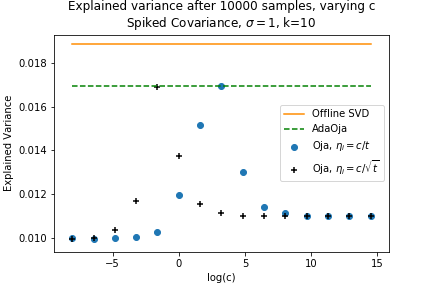}
    \caption{Spiked Covariance model for noise $\sigma=0.1, k \in \{1, 5, 10\}, B=10, n = 10000, d = 1000$, plot of final explained variance for Oja's method with learning rate $\eta_t = \frac{c}{t}, \frac{c}{\sqrt{t}}, c = 5^{i} \ \forall i \in \{-5, -4, \cdots, 10\}$.}
\end{figure}

\section{State-of-the-Art comparison for Synthetic Data}
\label{App:SotA comparison}
\begin{figure}[H]
    \centering
    \includegraphics[scale=.5]{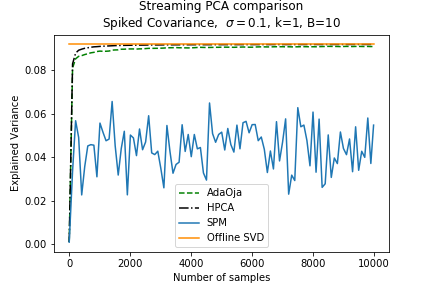}
    \includegraphics[scale=.5]{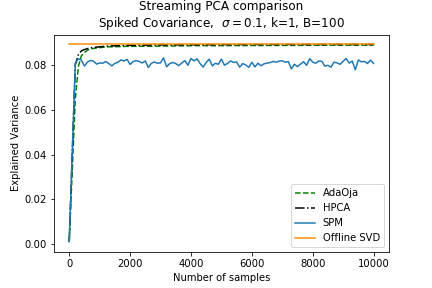}
    \includegraphics[scale=.5]{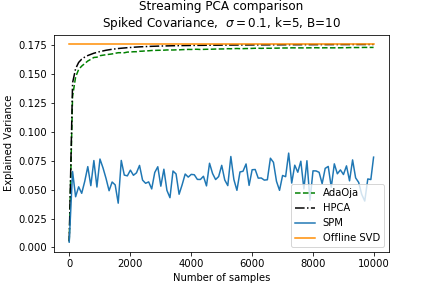}
    \includegraphics[scale=.5]{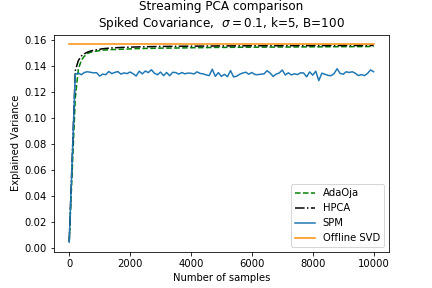}
    \includegraphics[scale=.5]{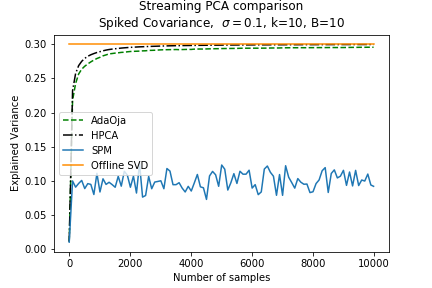}
    \includegraphics[scale=.5]{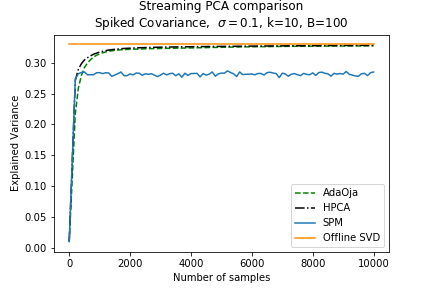}
    \caption{Explained variance plotted against the sample number for synthetic, spiked covariance data with noise $\sigma=0.01$. Here $k \in \{1, 5, 10\}, B \in \{10, 100\}, n=10000, d=1000$. The final explained variance computed in the offline setting is given here as a reference.}
\end{figure}

\begin{figure}[H]
    \centering
    \includegraphics[scale=.5]{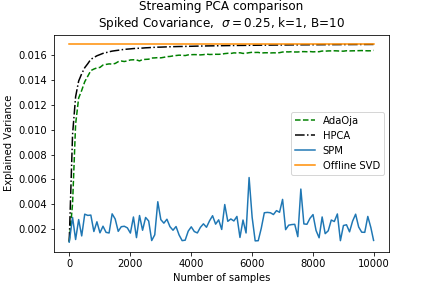}
    \includegraphics[scale=.5]{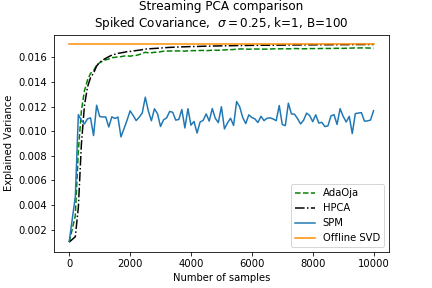}
    \includegraphics[scale=.5]{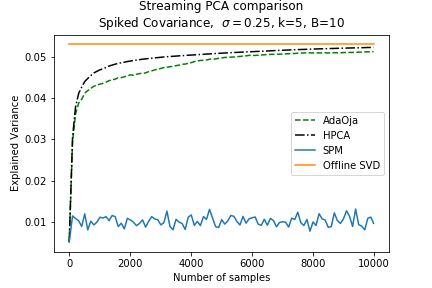}
    \includegraphics[scale=.5]{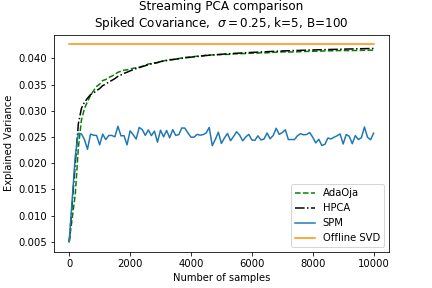}
    \includegraphics[scale=.5]{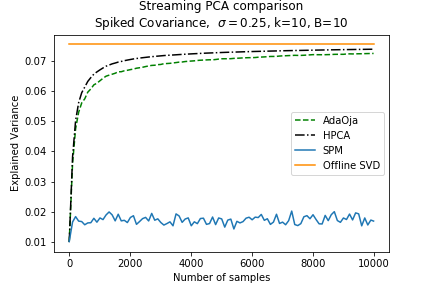}
    \includegraphics[scale=.5]{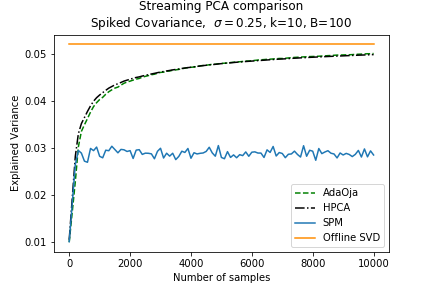}
    \caption{Explained variance plotted against the sample number for synthetic, spiked covariance data with noise $\sigma=0.25$. Here $k \in \{1, 5, 10\}, B \in \{10, 100\}, n=10000, d=1000$.}
\end{figure}

\begin{figure}[H]
    \centering
    \includegraphics[scale=.5]{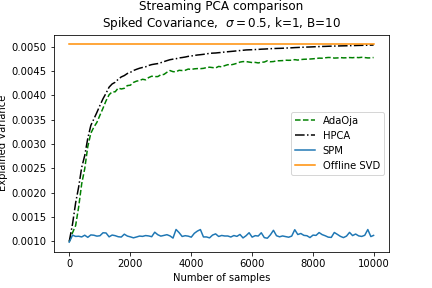}
    \includegraphics[scale=.5]{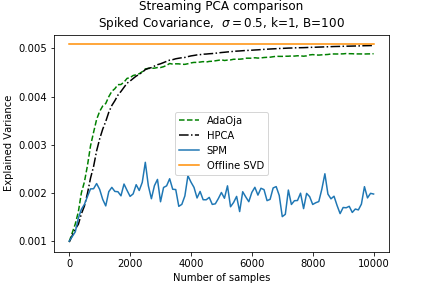}
    \includegraphics[scale=.5]{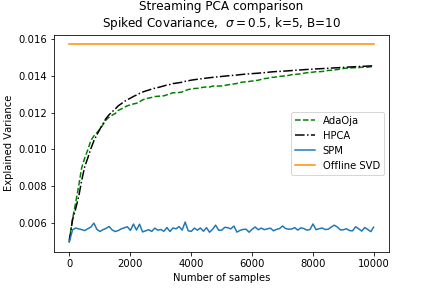}
    \includegraphics[scale=.5]{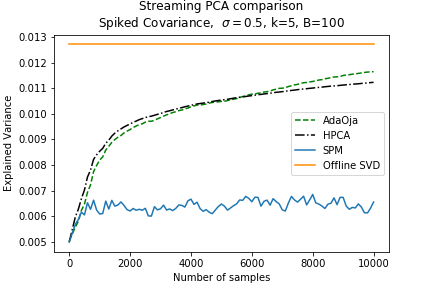}
    \includegraphics[scale=.5]{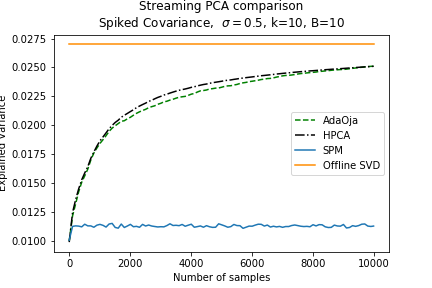}
    \includegraphics[scale=.5]{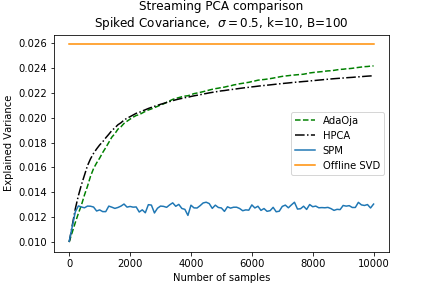}
    \caption{Explained variance plotted against the sample number for synthetic, spiked covariance data with noise $\sigma=0.5$. Here $k \in \{1, 5, 10\}, B \in \{10, 100\}, n=10000, d=1000$}
\end{figure}

\begin{figure}[H]
    \centering
    \includegraphics[scale=.5]{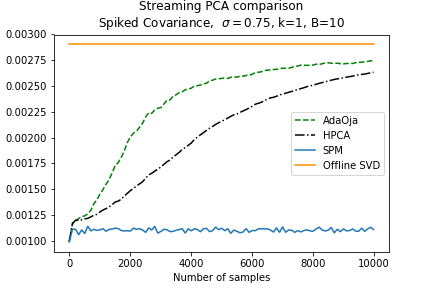}
    \includegraphics[scale=.5]{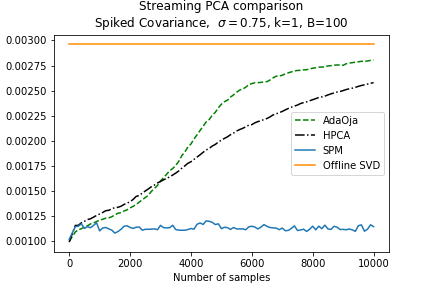}
    \includegraphics[scale=.5]{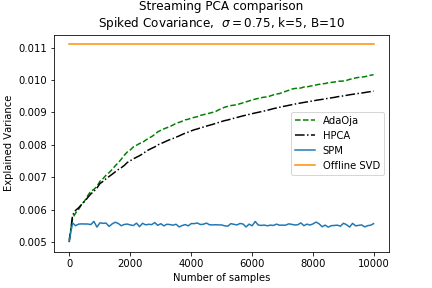}
    \includegraphics[scale=.5]{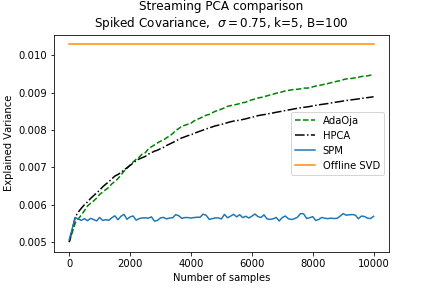}
    \includegraphics[scale=.5]{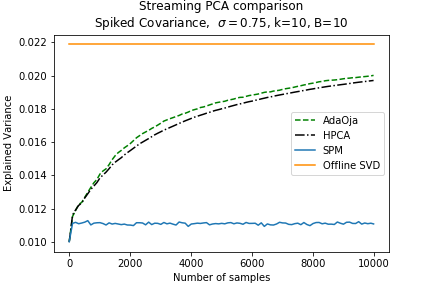}
    \includegraphics[scale=.5]{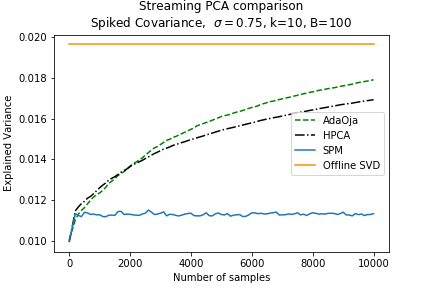}
    \caption{Explained variance plotted against the sample number for synthetic, spiked covariance data with noise $\sigma=0.75$. Here $k \in \{1, 5, 10\}, B \in \{10, 100\}, n=10000, d=1000$}
\end{figure}

\begin{figure}[H]
    \centering
    \includegraphics[scale=.5]{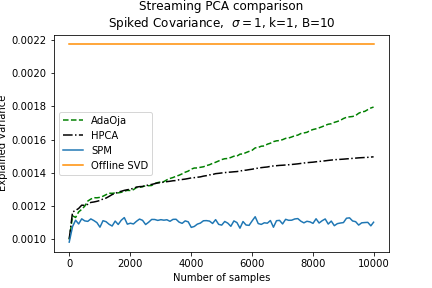}
    \includegraphics[scale=.5]{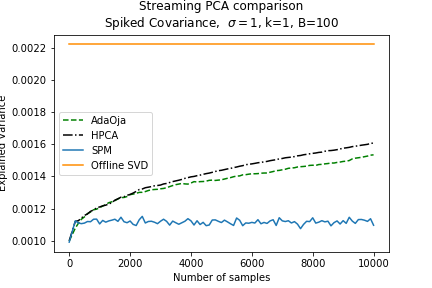}
    \includegraphics[scale=.5]{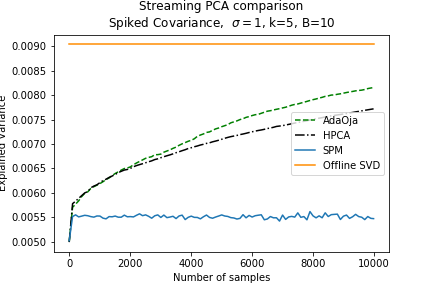}
    \includegraphics[scale=.5]{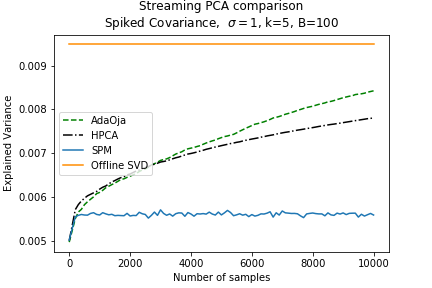}
    \includegraphics[scale=.5]{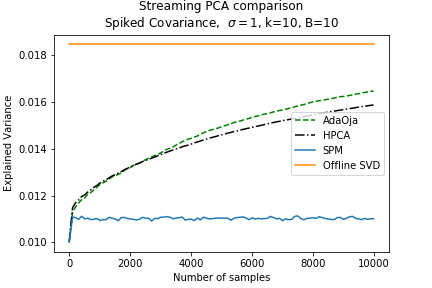}
    \includegraphics[scale=.5]{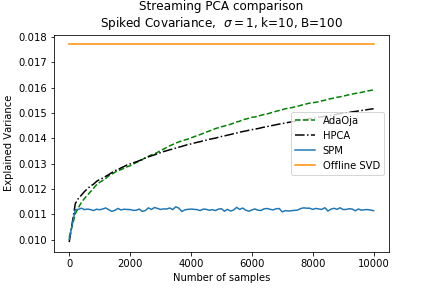}
    \caption{Explained variance plotted against the sample number for synthetic, spiked covariance data with noise $\sigma=1.0$. Here $k \in \{1, 5, 10\}, B \in \{10, 100\}, n=10000, d=1000$}
\end{figure}
\end{document}